\documentclass{article}

\usepackage{arxiv}
\usepackage{bm}
\usepackage[utf8]{inputenc} % allow utf-8 input
\usepackage[T1]{fontenc}    % use 8-bit T1 fonts
\usepackage{hyperref}       % hyperlinks
\usepackage{url}            % simple URL typesetting
\usepackage{booktabs}       % professional-quality tables
\usepackage{amsfonts}       % blackboard math symbols
\usepackage{nicefrac}       % compact symbols for 1/2, etc.
\usepackage{microtype}      % microtypography
\usepackage{lipsum}
\usepackage{adjustbox}
\usepackage{amsmath,amsthm,amssymb,amsfonts} % this for /equation* (without numbering)
\usepackage{algorithm}
\usepackage[noend]{algpseudocode}
\usepackage{soul}
\usepackage{booktabs,caption}
\usepackage[]{hyperref}
\hypersetup{
  colorlinks   = true, %Colours links instead of ugly boxes
  urlcolor     = blue, %Colour for external hyperlinks
  linkcolor    = blue, %Colour of internal links
  citecolor   = blue %Colour of citations
}

\title{Improving Intelligence of Evolutionary Algorithms Using Experience Share and Replay}

\author{
  Majdi I. Radaideh\thanks{Peer-review in progress} \\
  Department of Nuclear Science and Engineering,\\
  Massachusetts Institute of Technology,\\
  Cambridge, MA 02139, United States. \\
  \texttt{radaideh@mit.edu} \\
   \And
 Koroush Shirvan \\
  Department of Nuclear Science and Engineering,\\
  Massachusetts Institute of Technology,\\
  Cambridge, MA 02139, United States. \\
  \texttt{kshirvan@mit.edu} \\
}

\begin{document}
\maketitle

\begin{abstract}
We propose PESA, a novel approach combining Particle Swarm Optimisation (\textbf{P}SO), Evolution Strategy (\textbf{E}S), and Simulated Annealing (\textbf{S}A) in a hybrid \textbf{A}lgorithm, inspired from reinforcement learning. PESA hybridizes the three algorithms by storing their solutions in a shared replay memory. Next, PESA applies prioritized replay to redistribute data between the three algorithms in frequent form based on their fitness and priority values, which significantly enhances sample diversity and algorithm exploration. Additionally, greedy replay is used implicitly within SA to improve PESA exploitation close to the end of evolution. The validation against 12 high-dimensional continuous benchmark functions shows superior performance by PESA against standalone ES, PSO, and SA, under similar initial starting points, hyperparameters, and number of generations. PESA shows much better exploration behaviour, faster convergence, and ability to find the global optima compared to its standalone counterparts. Given the promising performance, PESA can offer an efficient optimisation option, especially after it goes through additional multiprocessing improvements to handle complex and expensive fitness functions.  

\end{abstract}

% keywords can be removed
\keywords{Evolutionary Computation \and Continuous Optimisation \and Simulated Annealing \and Experience Replay \and Hybrid Algorithms}

\section{Introduction}

Solving optimisation problems is at the heart of every scientific discipline to improve our understanding and interpretation of scientific findings. Evolution-based and swarm intelligence search and optimisation have been in remarkable growth over the years to tackle complex problems ranging from scientific research \cite{abraham2006swarm} to industry \cite{sanchez2012industrial} and commerce \cite{freitas2002review}. Hybridizing evolutionary, swarm, and annealing algorithms \cite{grosan2007hybrid} (the focus of this work) is an active area of research, since usually hybrid algorithms can offer several advantages over standalone algorithms in terms of stability, search speed, and exploration capabilities, where these advantages highlight the goals of these hybrid studies. Few examples of successful demonstration of hybrid algorithms in different domains are: (1) genetic algorithm (GA) and particle swarm optimisation (PSO) \cite{kao2008hybrid}, (2) GA and Simulated Annealing (SA) \cite{chen2009hybrid}, (3) GA and SA \cite{ma2014hybrid}, (4) PSO and tabu search \cite{shen2008hybrid}, (5) PSO and evolution strategies (ES) \cite{jamasb2019novel}, and many more. For the authors' interests on nuclear power plant design, evolutionary, swarm, and annealing algorithms have been proposed in several research studies in standalone and hybrid forms to optimise nuclear fuel assemblies and cores of light water reactors \cite{zameer2014core,rogers2009optimization,de2009particle}, to reduce fuel costs and improve nuclear reactor safety \cite{kropaczek1991core}. 

As improving capabilities of evolutionary and stochastic algorithms in solving optimisation problems is always a research target due to their widely usage, we propose a new algorithm called PESA (\textbf{P}SO, \textbf{E}S, and \textbf{S}A \textbf{A}lgorithm). PESA hybridizes three known optimisation techniques by exchanging their search data on-the-fly, storing all their solutions in a buffer, and replaying them frequently based on their importance. The concept of experience replay was first introduced in deep reinforcement learning (RL) \cite{mnih2015human,schaul2015prioritized} to improve agent learning by replaying relevant state/action pairs weighted by their temporal difference and reward values. Accordingly, we introduce experience replay into evolutionary algorithms to determine whether this concept would improve the performance. First, we hybridize PSO, ES, and SA to create a replay memory with diverse samples by taking the search advantages of each individual algorithm. Next, two modes of experience replay are applied: (1) prioritized replay to enhance exploration capabilities of PESA, and (2) ``backdoor'' greedy replay to improve algorithm exploitation, such that once the search is close to end, PESA prioritizes its main experience. To enhance search speed, PSO, ES, and SA are parallelized during search such that the three algorithms can run simultaneously to collect experiences, before updating the memory and executing the replay. We benchmark PESA against its standalone components: PSO, ES, and SA to show its promise. Variety of commonly used continuous benchmark functions are utilized in this paper, which also have a high dimensional nature to represent realistic optimisation problems. We evaluate PESA performance by its ability to find a known global optima, exploration/exploitation capabilities, and computational efficiency.              

\section{Methodology}
\label{sec:method}

\begin{figure}[h] 
 \centering
  \includegraphics[width=\textwidth]{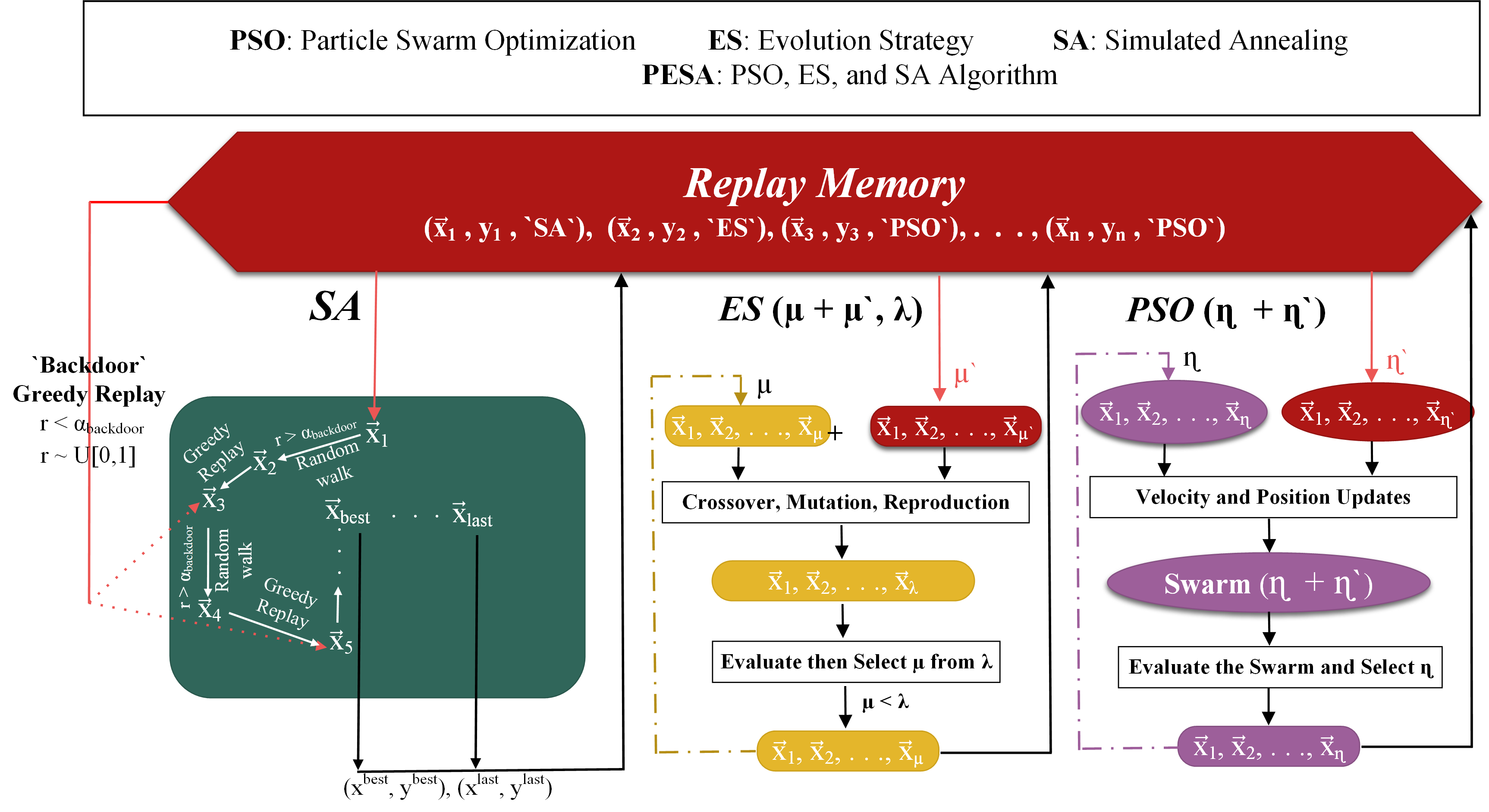}
  \caption{PESA algorithm workflow. Red arrows represent memory feedback, $\alpha_{backdoor}$ is the probability to use backdoor greedy replay in SA. $\eta$ and $\eta'$ are the number of PSO particles surviving from previous generation and from the memory, respectively. $\mu$ and $\mu'$ are the number of ES individuals surviving from previous generation and from the memory, respectively. $\lambda$ is the full size of ES offspring. All hyperparameters are defined in detail later in this section}
  \label{fig:pesa}
\end{figure}

The workflow of PESA is sketched in Figure \ref{fig:pesa}, which shows how the algorithms are connected to each other through the replay memory. The choices of PSO, ES, and SA are not arbitrary. PSO \cite{kennedy1995particle} is known to excel in continuous optimisation, ES \cite{beyer2002evolution} (which originated from the genetic algorithms) performs well in both continuous/discrete spaces, while SA \cite{kirkpatrick1983optimization} is introduced to ensure exploration due to its stochastic nature, and more importantly to drive PESA exploitation to the best solution found, as will be described in the next section.    

\subsection{Evolutionary, Swarm, and Annealing Computation}
\label{sec:evol}

Evolution strategies (ES) \cite{beyer2002evolution} are inspired by the theory of natural selection. In this work, the well-known $(\mu, \lambda)$ strategy is used, which is indeed similar to the continuous GA in terms of the operators (e.g., crossover, mutation, and reproduction). However, the mutation strength of each gene/attribute in the individual ($\vec{x}$) is not constant, but learned during the evolution. Accordingly, each individual in the population carries a strategy vector of same size as ($\vec{x}$), where the strategy vector is adapted during the evolution, and controls the mutation strength of each attribute. We adopt log-normal adaptation for the strategy vector (see \cite{beyer2002evolution}), where the min and max strategies are bounded between $1/n$ and 0.5, respectively, as suggested by the literature, where $n$ is the size of $\vec{x}$. On the population level, the crossover operation selects two random individuals for mating with probability $CX$, where we use the classical two-point crossover in this work. After crossover, some of the individuals may be selected for mutation with a small probability $MUT$. Notice that for $(\mu, \lambda)$, both $MUT$ and $CX$ on the population level remain fixed, unlike the internal mutation strength for each individual. After fitness evaluation of the population, the selected individuals $\mu$ are passed to the next generation, where these $\mu$ individuals participate in generating the next offspring of size $\lambda$ (i.e., $\lambda \geq \mu$).   

Particle swarm optimisation (PSO) \cite{kennedy1995particle} is inspired by the movement of organisms in bird or fish flocks. Each particle in the swarm experiences position update ($x^{t+1}_i = x^t_i + v^{t+1}_i$), where $i$ is the attribute index and $v$ is the velocity value for that attribute. We implement the constriction approach by Clerc and Kennedy \cite{clerc2002particle} for velocity update, which can be expressed as follows
\begin{align} 
v^{t+1}_i &= K[v^t_i + c_1r_1(pbest^t_i - x^t_i) + c_2r_2(gbest^t - x^t_i)] \\
K &=  \frac{2}{|2- \phi - \sqrt{\phi^2-4\phi}|} \\
\phi &= c_1 + c_2, \quad \phi > 4
\end{align}
where $c_1$, $c_2$ are the cognitive and social speed constants, respectively, $r_1$, $r_2$ are uniform random numbers between [0,1], and $pbest$, $gbest$ are the local best position for each particle and global best position of the swarm, respectively. Lastly, $K$ is the constriction coefficient introduced to balance PSO exploration/exploitation and improve stability. Typically, when $c_1 = c_2 = 2.05$, then $K=0.73$. Another advantage of using constriction is it exempts us from using velocity clamping; therefore there is no need to specify minimum and maximum velocities, which reduces PSO hyperparameters, and PESA by definition. The number of particles in the swarm is given by $\eta$.   

Simulated Annealing (SA) \cite{kirkpatrick1983optimization} is inspired from the concept of annealing in physics to reduce defects in crystals through heating followed by progressive cooling. In optimisation, SA combines high climbing and pure random-walk to help us find an optimum solution through implementing five basic steps: (1) generate a candidate solution, (2) evaluate candidate fitness, (3) generate a random neighbor solution and calculate its fitness, (4) compare the old and new fitness evaluations (i.e., $\Delta E$ increment), if better continue with the new solution, if worse, accept the old solution with probability $\alpha=exp^{-\Delta E/T}$, where T is the annealing temperature, (5) repeat steps 3-4 until convergence. Temperature is annealed between $T_{max}$ and $T_{min}$ over the annealing period, where the fast annealing schedule is adopted in this work
\begin{equation}
\label{eq:fast}
   T = T_{max}\cdot exp\bigg[\frac{-log(T_{max}/T_{min})N}{N_{steps}}\bigg],
\end{equation}
where $N$ is the current annealing step, which builds up from 1 to $N_{steps}$. New SA candidates are generated using random-walk with a small probability $\chi$, where each input attribute is subjected to perturbation once $rand \sim U[0,1] < \chi$ is satisfied. Small value of $\chi$ between 0.05-0.15 seemed to yield better performance in this work. 

\subsection{Experience Replay}
\label{sec:er}

Two modes of experience replay are relevant to this work: (1) greedy and (2) prioritized replay. Greedy replay sorts the memory from min to max (assuming a minimization problem), and replays the sample(s) with lowest fitness every time. Intuitively, replaying always the best samples could lead to improvement early on, but will restrict the diversity of the search, and likely to end with premature convergence (i.e., methods converge to a local optima). Therefore, greedy replay in PESA is used in a special form with SA described later. Prioritized replay balances between uniform and greedy sampling through using uniform replay early on, allowing PESA to explore more at first and exploit at the end. Probability of replaying sample $i$ can be defined as follows: 
\begin{equation}
   \label{eq:prior1}
    P_i = \frac{p_i^\alpha}{\sum_{d=1}^D p_d^\alpha}
\end{equation}
where $D$ is the current memory size and exponent $\alpha$ is the priority coefficient $\in [0,1]$, explaining how much prioritization is used. $\alpha = 0$ corresponds to uniform replay (i.e., all samples have equal weights), while $\alpha =1$ corresponds to the fully prioritized replay. Notice here we describe $\alpha =1$ as ``fully prioritized`` replay rather than greedy replay (i.e., min operator). When $\alpha =1$, the best samples (lowest fitness) are ``more likely'' to be replayed due to their larger weight, but in greedy replay these quality samples are ``100\%'' replayed (because of applying the $min$ operator). Next, $p_i$ refers to the priority of sample $i$, which can take different forms. For our case, we choose a prioritization based on sample rank:
\begin{equation}
   \label{eq:prior2}
    p_i = \frac{1}{rank(i)}
\end{equation}
where $rank(i)$ is the sample rank based on fitness value when the memory is sorted from minimum fitness to maximum. After sorting, sample 1 has $p_i = 1$, sample 2 has $p_i = 0.5$, and so on. According to \cite{schaul2015prioritized}, rank-based prioritization is robust since it is insensitive to outliers. Now, to balance between exploration/exploitation in prioritized replay, we start PESA evolution with small value of $\alpha = \alpha_{init} = 0.01$ (more exploration), and then gradually increasing until $\alpha =\alpha_{end} = 1.0$ (more exploitation) by the end of evolution. After a generation $k$, priorities are calculated by Eq.\eqref{eq:prior2}, sampling probability is determined by Eq.\eqref{eq:prior1}, and replay is performed by sampling from a non-uniform distribution with probabilities [$P_1, P_2, ..., P_D$]. Notice that redundant samples (if any) are removed from the memory before re-sampling to avoid biasing the replay toward certain samples. 

\begin{algorithm}[!h]
  \small
    \caption{Evolution Strategy ($\mu + \mu', \lambda$) in PESA}
    \label{alg:es}
    \begin{algorithmic}[1]
    \State \textbullet Given ES hyperparameters: $\mu$, $\mu'$, $\lambda$, $MUT$, $CX$
    \For{a generation $k$}
        \State \textbullet Apply crossover operator to the population ($\mu + \mu'$) with probability $CX$
        \State \textbullet Apply mutation operator to the population ($\mu + \mu'$) with probability $MUT$
        \If {$MUT + CX < 1$}
            \State \textbullet Apply reproduction to the population ($\mu + \mu'$) with probability $1 - MUT - CX$
        \EndIf
        \State \textbullet Generate final offspring $\lambda$ from the population ($\mu + \mu'$)
        \State \textbullet \textit Evaluate fitness of the offspring
        \State \textbullet Select $\mu$ individuals with best fitness for next generation
    \EndFor
    \State \textbullet Return the selected individuals $\bm{\mu} = \{(\vec{x}_1, y_1), (\vec{x}_2, y_2),...,(\vec{x}_\mu, y_\mu)\}$
    \end{algorithmic}
\end{algorithm}

As in Figure \ref{fig:pesa}, prioritized replay is used with all three algorithms at the beginning of each generation. For ES, the replay memory provides $\mu'$ individuals at the beginning of every generation, which mix with original $\mu$ individuals (from previous generation) using crossover, mutation, and reproduction operations (see Figure \ref{fig:pesa}). Therefore, the ($\mu,\lambda$) algorithm becomes ($\mu + \mu',\lambda$) for PESA, as $\mu'$ individuals are mixed with $\mu$ before forming the next $\lambda$ offspring. The ES algorithm in PESA is given in Algorithm \ref{alg:es}.

The prioritized replay for PSO is similar to ES, where the swarm particles at every generation constitute from $\eta$ particles of the previous generation and $\eta'$ particles from the memory, before going through velocity and position updates as described in section \ref{sec:evol}. In both cases (ES, PSO), the values of $\mu'$ and $\eta'$ can be tuned for better performance. The PSO algorithm in PESA is given in Algorithm \ref{alg:pso}.  

\begin{algorithm}[!h]
  \small
    \caption{Particle Swarm Optimisation ($\eta + \eta'$) in PESA}
    \label{alg:pso}
    \begin{algorithmic}[1]
    \State \textbullet Given PSO hyperparameters: $\eta$, $\eta'$, $c_1$, $c_2$
    \For{a generation $k$}
        \State \textbullet Update velocity of swarm particles ($\eta + \eta'$) with constriction coefficient
        \State \textbullet Generate new swarm by updating all particle positions ($\eta + \eta'$) 
        \State \textbullet Evaluate fitness of the swarm
        \State \textbullet Select $\eta$ particles with best fitness for next generation
    \EndFor
    \State \textbullet Return the selected particles $\bm{\eta} = \{(\vec{x}_1, y_1), (\vec{x}_2, y_2),...,(\vec{x}_\eta, y_\eta)\}$
    \end{algorithmic}
\end{algorithm}

For SA, prioritized replay is used to make an initial guess for the chain before starting a new annealing generation. Given SA is running in series in this work, a single chain is used. Once the generation is done, SA updates the memory by the last pair ($\vec{x}^{last},y^{last}$) and best pair ($\vec{x}^{best},y^{best}$) observed by the chain in that generation. The SA algorithm in PESA is given in Algorithm \ref{alg:sa}.

\begin{algorithm}[!h]
  \small
    \caption{Simulated Annealing in PESA}
    \label{alg:sa}
    \begin{algorithmic}[1]
    \State \textbullet Given SA hyperparameters: $T_{max}$, $T_{min}$, $\alpha_{backdoor}$
    \State \textbullet Draw initial state from the memory, $\theta' = (\vec{x}_0, y_0)$ 
    \State \textbullet Set $\vec{x}_{prev} \xleftarrow{} \vec{x}_0$, $E_{prev} \xleftarrow{} y_0$
       \For{a generation $k$}
            \If {$rand \sim$ U[0,1] $< \alpha_{backdoor}$}
                \State \textbullet Draw the best sample from the memory $(\vec{x}',y')$
                \State \textbullet Set $\vec{x}\xleftarrow{} \vec{x}'$, $E \xleftarrow{} y'$ 
            \Else
                \State \textbullet Perform a random walk as next chain state $(\vec{x})$
                \State \textbullet Evaluate fitness $E$ for the new state
            \EndIf
            \State \textbullet Calculate $\Delta E = E - E_{prev}$ 
            \If {$\Delta E < 0$ OR $exp(-\Delta E/T) > rand \sim$ U[0,1]}
                \State \textbullet Accept the candidate 
                \State \textbullet Set $E_{prev} \xleftarrow{} E$, $\vec{x}_{prev} \xleftarrow{} \vec{x}$  
            \Else
                \State \textbullet Reject the candidate and restore previous state
                \State \textbullet Set $E_{prev} \xleftarrow{} E_{prev}$, $x_{prev} \xleftarrow{} x_{prev}$   
            \EndIf
            \State \textbullet Anneal $T$ between $T_{max}$ and $T_{min}$ 
        \EndFor
    \State \textbullet Return the last chain state ($\vec{x}^{last},y^{last}$) and best state ($\vec{x}^{best},y^{best}$)
\end{algorithmic}
\end{algorithm}

The backdoor greedy replay for SA in Figure \ref{fig:pesa} and Algorithm \ref{alg:sa} has two main benefits: (1) ensuring PESA exploitation at the end of evolution and (2) providing more guidance to SA that relies extensively on random-walk. Unlike prioritized replays that occur explicitly at the beginning of every generation of all algorithms including SA, backdoor replay obtained this name since it occurs implicitly during SA generation, by giving the SA chain a choice between its regular random-walk or the best quality sample in the memory with probability $rand \sim U[0,1] < \alpha_{backdoor}$. Since SA tends to accept low quality solutions early on, but tightens the acceptance criteria at the end by rejecting low quality solutions, this means SA will implicitly drive PESA to always converge to the best solution in the memory once the evolution is close to end. Second, as SA lacks the learning capabilities of PSO and ES (velocity update, crossover, etc.), the backdoor replay will frequently correct the SA chain to focus the search in relevant space regions as found by her sisters (PSO, ES) so far during the evolution. We typically recommend small value for $\alpha_{backdoor}$ (i.e., $<$ 0.15), since obviously large values lead to repetitive greedy replays, which in turn cause a bias in the chain, eventually leading to premature convergence in SA.  

\subsection{PESA Algorithm}

By combining all parts presented in sections \ref{sec:evol}-\ref{sec:er}, PESA algorithm can be constructed as given in Algorithm \ref{alg:pesa}. The flow of PESA can be summarised in three main phases:
\begin{enumerate}
    \item Warmup (lines 1-4 in Algorithm \ref{alg:pesa}): Hyperparameters of all individual algorithms (ES, PSO, SA) are specified. The memory is initialized with a few warmup samples ($N_{warmup}$) and maximum capacity ($D_{max}$).
    \item Evolution (lines 8-17 in Algorithm \ref{alg:pesa}): The three algorithms, ES, PSO, and SA are executed in parallel according to Algorithm \ref{alg:es}, Algorithm \ref{alg:pso}, and Algorithm \ref{alg:sa}, respectively. \textit{Each individual algorithm runs its iterations in serial}. 
    \item Memory management (lines 6-7, 18-20 in Algorithm \ref{alg:pesa}): This phase involves updating the memory with new samples, calculating and updating sample priority, annealing the prioritization coefficient ($\alpha$), and cleaning the memory from duplicates. 
\end{enumerate}
    
\begin{algorithm}[!h]
  \small
    \caption{PESA Algorithm with Prioritized Replay}
    \label{alg:pesa}
    \begin{algorithmic}[1]
     \State \textbullet Set hyperparameters of ES, SA, PSO
     \State \textbullet Set replay parameters: $\alpha_{backdoor}$, $\alpha_{init}$, $\alpha_{end}$
     \State \textbullet Construct the replay memory with size $D_{max}$ and initialize with warmup samples $N_{warmup}$  
     \State \textbullet Set $\alpha \xleftarrow{} \alpha_{init}$
     \For{GEN $i = 1$ to $N_{gen}$}
        \State \textbullet Calculate $p_i = 1/rank(i)$ and priorities $P(i) = p_i^\alpha/\sum_d p_d^\alpha$
        \State \textbullet With probabilities $P(i)$, draw samples $\bm{\mu'}$, $\bm{\eta'}$, and $\bm{\theta'}$
        \For{\textit{Parallel} Process 1: ES}
            \State \textbullet Given $\bm{\mu'} = \{(\vec{x}_1, y_1),...,(\vec{x}_\mu, y_{\mu'})\}$, run ES Algorithm \ref{alg:es}
            \State \textbullet Obtain ES population $\bm{\mu} = \{(\vec{x}_1, y_1), (\vec{x}_2, y_2),...,(\vec{x}_\mu, y_\mu)\}$
        \EndFor

        \For{\textit{Parallel}  Process 2: PSO}
            \State \textbullet Given $\bm{\eta'} = \{(\vec{x}_1, y_1),...,(\vec{x}_\eta, y_{\eta'})\}$, run PSO Algorithm \ref{alg:pso}
            \State \textbullet Obtain PSO population $\bm{\eta} = \{(\vec{x}_1, y_1), (\vec{x}_2, y_2),...,(\vec{x}_\eta, y_\eta)\}$
        \EndFor
        
        \For {\textit{Parallel} Process 3: SA}
            \State \textbullet Given $\bm{\theta'} = \{(\vec{x}', y')\}$, run SA Algorithm \ref{alg:sa} 
            %\State \textbullet Obtain SA population $\bm{\theta} = \{(\vec{x}_1^{last}, y_1^{last}),...,(\vec{x}_\theta^{last}, y_\theta^{last})\}$
            
            %\State \textbullet Obtain SA best population $\bm{\theta}^{best} = \{(\vec{x}_1^{best}, y_1^{best}),...,(\vec{x}_\theta^{best}, y_\theta^{best})\}$.
            \State \textbullet Obtain SA population $\bm{\theta} = \{(\vec{x}^{last}, y^{last})\}$
            
            \State \textbullet Obtain SA best population $\bm{\theta}^{best} = \{(\vec{x}^{best}, y^{best})\}$
        \EndFor 
        
        \State \textbullet Update the memory with samples $\bm{\mu}, \bm{\eta}$, $\bm{\theta}$, and $\bm{\theta}^{best}$
        \State \textbullet Remove duplicated samples from the memory
        \State \textbullet Anneal $\alpha$ between $\alpha_{init}$, $\alpha_{end}$
    \EndFor 
\end{algorithmic}
\end{algorithm}

As can be noticed from Algorithm \ref{alg:pesa}, algorithm-based parallelism can be observed in PESA (see steps 8, 11, and 14 in Algorithm \ref{alg:pesa}), which involves running the three algorithms simultaneously. This feature exists in the algorathim, but not activated in this work, since the functions investigated are too fast-to-evaluate. Internal parallelization of each algorithm is left for future, since again the benchmark functions tested in this work are cheap-to-evaluate, so multiprocessing of each algorithm is also not advantageous.

With an effort to minimize the number of hyperparameters in PESA, we assume similar size of generation across algorithms. For example, lets assume a generation of 60 individuals assigned for each algorithm. In this case, the ES population has size $\lambda=60$, PSO swarm has 60 particles, SA chain has size of $C_{size}=60$ steps. Although previous assumptions do not necessarily guarantee the most optimal performance, the authors believe that such symmetry in PESA has two main advantages. First, the burden of hyperparameter optimisation is significantly reduced. Second, algorithm dynamics is improved as all internal algorithms will finish their generation roughly same time, which allows them to stay up to date with the memory. This is clearly under the assumption that fitness evaluation cost ($y$) is roughly the same regardless of the input value ($\vec{x}$).

\section{Numerical Tests}
\label{sec:tests}

The mathematical forms of selected benchmark functions are given in Table \ref{tab:funcs}, which are all well-known optimisation benchmarks \cite{jamil2013literature}. All functions have $n$-dimensions nature, where here we select $n=50$ to represent a more high-dimensional problem. All functions have a known global minima at $f(\vec{x}) = 0$, except for the Ridge and Exponential functions, which have their global minima at -5 and -1, respectively.    

% Table generated by Excel2LaTeX from sheet 'method_compare'
\begin{table}[htbp]
  \centering
  \footnotesize
  \caption{List of continuous benchmark functions with $n$-dimensional space}
 \begin{adjustbox}{max width=\textwidth}
    \begin{tabular}{llll}
    \toprule
    Name & Formula & $\vec{x}$ Range & Global Optima \\
    \midrule
    (1) Cigar   & $ f(\vec{x}) = x_0^2 + 10^6\sum_{i=1}^n\,x_i^2$ & $[-10,10]^n$ & $\vec{x}^* = \vec{0}, f(\vec{x}^*) = 0$   \\
    (2) Sphere   & $ f(\vec{x}) = \sum_{i=1}^n x_i^2$ & $[-100,100]^n$ & $\vec{x}^* = \vec{0}, f(\vec{x}^*) = 0$   \\
    (3) Ridge   & $ f(\vec{x}) = x_1 + (\sum_{i=2}^{n}x_i^2)^{0.5}$ & $[-5,5]^n$ & $\vec{x}^* = [-5,0, ..., 0], f(\vec{x}^*) = -5$   \\
    (4) Ackley    & $    f(\vec{x}) = 20-20exp\Big(-0.2\sqrt{\frac{1}{n}\sum_{i=1}^{n}x_i^2}\Big)-exp\Big(\frac{1}{n}\sum_{i=1}^{n}cos(2\pi x_i)\Big) + e$ & $[-32,32]^n$ & $\vec{x}^* = \vec{0}, f(\vec{x}^*) = 0$   \\
    (5) Bohachevsky    & $f(\vec{x}) = \sum_{i=1}^{n-1}(x_i^2 + 2x_{i+1}^2 - 0.3\cos(3\pi x_i) - 0.4\cos(4\pi x_{i+1}) + 0.7)$ & $[-100,100]^n$ & $\vec{x}^* = \vec{0}, f(\vec{x}^*) = 0$   \\
    (6) Griewank   & $ f(\vec{x}) = \frac{1}{4000}\sum_{i=1}^n\,x_i^2 - \prod_{i=1}^n\cos\left(\frac{x_i}{\sqrt{i}}\right) + 1$ & $[-600,600]^n$ & $\vec{x}^* = \vec{0}, f(\vec{x}^*) = 0$   \\
    % (7) Schaffer   & $ f(\vec{x}) = \sum_{i=1}^{n-1} (x_i^2+x_{i+1}^2)^{0.25} [\sin^2(50(x_i^2+x_{i+1}^2)^{0.1}) + 1.0]$ & $[-100,100]^n$ & $\vec{x}^* = \vec{0}, f(\vec{x}^*) = 0$   \\
    (7) Brown   & $ f(\vec{x}) = \sum_{i=1}^{n-1}(x_i^2)^{(x_{i+1}^{2}+1)}+(x_{i+1}^2)^{(x_{i}^{2}+1)}$ & $[-1,4]^n$ & $\vec{x}^* = \vec{0}, f(\vec{x}^*) = 0$   \\
    (8) Exponential   & $ f(\vec{x})=-exp(-0.5\sum_{i=1}^n{x_i^2})$ & $[-1,1]^n$ & $\vec{x}^* = \vec{0}, f(\vec{x}^*) = -1$   \\
    (9) Zakharov   & $ f(\vec{x})=\sum_{i=1}^n x_i^{2}+(\sum_{i=1}^n 0.5ix_i)^2 + (\sum_{i=1}^n 0.5ix_i)^4$ & $[-5,10]^n$ & $\vec{x}^* = \vec{0}, f(\vec{x}^*) = 0$   \\
    (10) Salomon   & $ f(\vec{x})=1-cos(2\pi\sqrt{\sum_{i=1}^{n}x_i^2})+0.1\sqrt{\sum_{i=1}^{n}x_i^2}$ & $[-100,100]^n$ & $\vec{x}^* = \vec{0}, f(\vec{x}^*) = 0$   \\
    (11) Quartic   & $ f(\vec{x})=\sum_{i=1}^{n}ix_i^4+\text{random}[0,1)$ & $[-1.28,1.28]^n$ & $\vec{x}^*=\vec{0}, f(\vec{x}^*) = 0 + \text{noise}$   \\
    (12) Levy   & $\begin{aligned} f(\vec{x}) & = sin^2(\pi w_1) + \sum_{i=1}^{n-1}(w_i-1)^2[1+10sin^2(\pi w_i+1)] \\ & + (w_n-1)^2[1+sin^2(2\pi w_n)], \quad w_i = 1 + (x_i-1)/4  \end{aligned}$  & $[-10,10]^n$ & $\vec{x}^* = \vec{1}, f(\vec{x}^*) = 0$   \\
    \bottomrule
    \end{tabular}%
   \end{adjustbox}
  \label{tab:funcs}%
\end{table}%

The hyperparameters selected to perform the tests are as follows: For ES ($CX=0.6, MUT=0.15, \lambda=60, \mu=30, \mu'=30$), for PSO ($c_1=2.05, c_2=2.05, \eta=30, \eta'=30$), for SA ($T_{max} = 10000, T_{min} = 1, \chi=0.1, C_{size} = 60$), while for replay parameters ($\alpha_{init}=0.01, \alpha_{end}=1.0, \alpha_{backdoor}=0.1$). The tests are executed with $N_{gen} = 100$ generations and $N_{warmup} = 500$ samples. The previous hyperparameters yielded satisfactory performance for all test functions. \textit{It is very important highlighting that the hyperparameters, initial starting points, and number of generations are preserved between PESA and the standalone algorithms to isolate their effects}. This means that any difference in performance comes purely from the ``experience share and replay'', the contribution of this work.   

The convergence of fitness results is plotted in Figure \ref{fig:pesa}, which compares PESA against the standalone algorithms given the prescribed hyperparameters. For brevity, the plotted results include the minimum fitness found in each generation, since it is our end goal, so the first generation point is not necessarily the initial guess in all algorithms. In addition, log-scale is used for some functions (Cigar, Sphere, Brown, etc.) to reflect a better scale, where we set $10^{-2}$ as a lower bound to represent the zero global minima. Obviously, the results clearly show PESA outperforming the standalone ES, PSO, and SA by a wide margin in terms of number of generations to reach global minima. First, in all benchmarks, PESA successfully found and converged to the global optima, while the standalone algorithms failed to do so in most cases. In addition, PESA tends to converge much faster to the feasible region, thanks to the collaborative environment that PESA creates. We can notice that PSO is the most competitive algorithm to PESA. This is expected, since PSO was developed for continuous optimization, which is the feature of all the considered benchmarks. However, PSO alone seems to converge slowly. Due to the high dimensionality, standalone SA struggles to resolve the search space or to converge to a relevant solution in all cases. Standalone ES performance in Figure \ref{fig:pesa} is bounded between PSO and SA in most cases, except for the Levy function at which ES outperforms PSO.

\begin{figure}[h] 
 \centering
  \includegraphics[width=\textwidth]{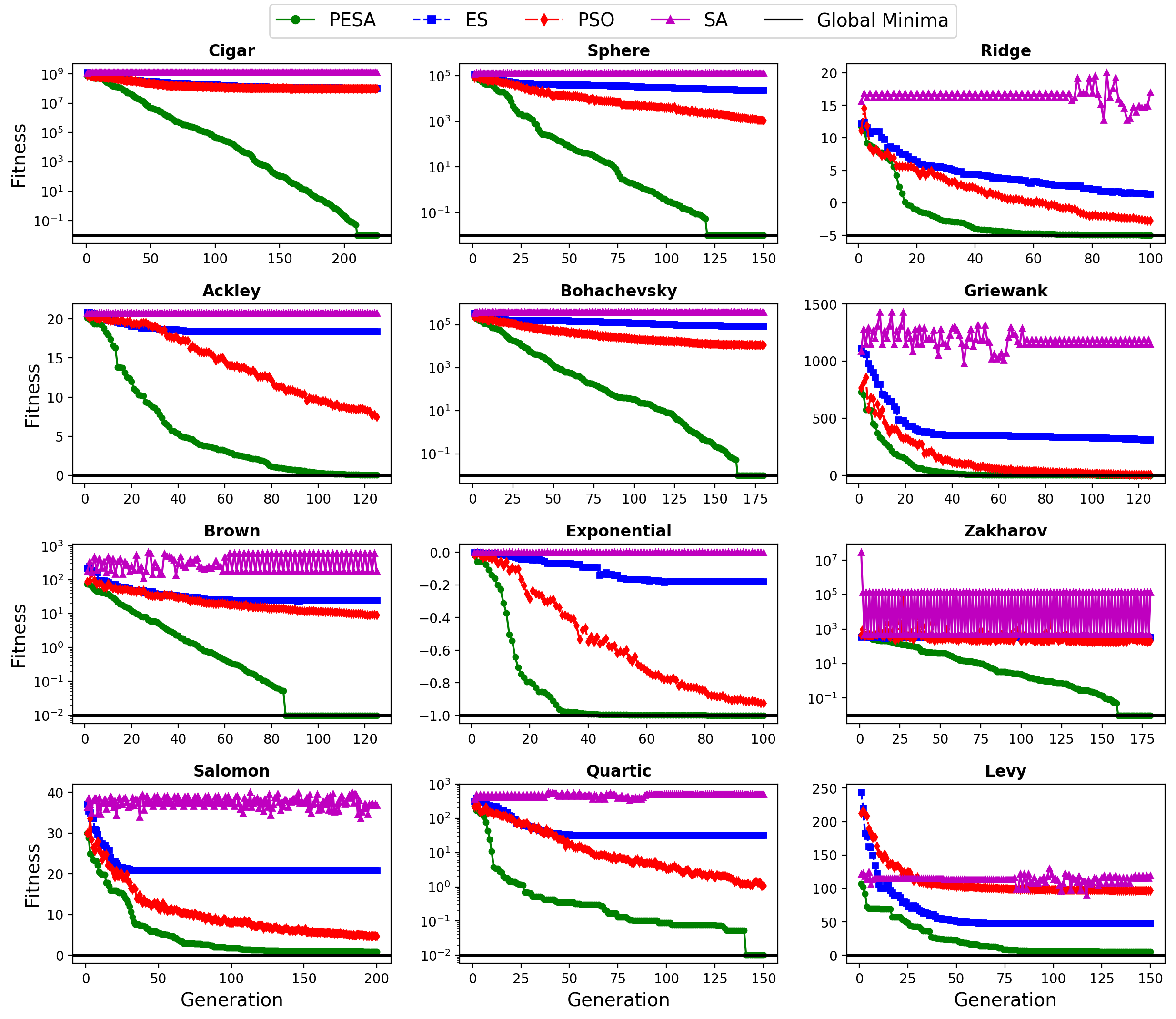}
  \caption{Comparison of fitness convergence of PESA against ES, PSO, and SA in \textbf{standalone} form for minimization of 12 continuous benchmarks, dimensionality $n=50$ (See Table \ref{tab:funcs})}
  \label{fig:bench}
\end{figure}

\begin{figure}[h] 
 \centering
  \includegraphics[width=0.55\textwidth]{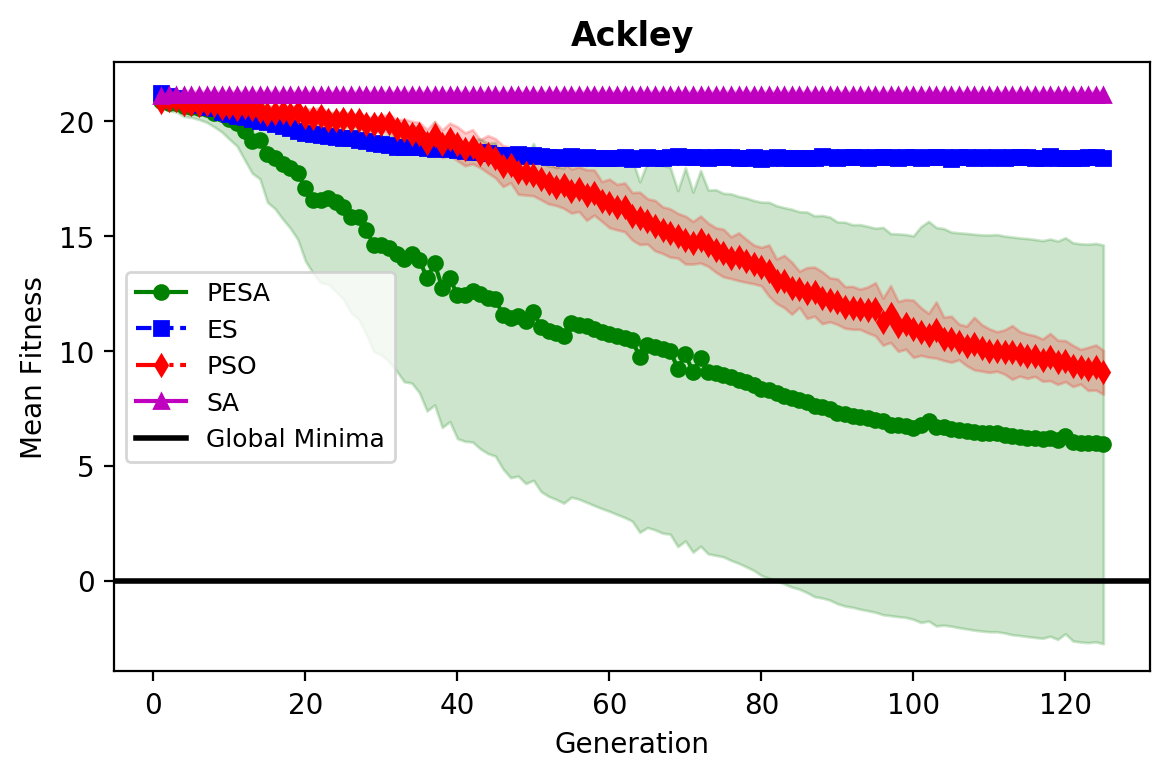}
  \caption{Convergence of fitness mean and standard deviation of PESA against \textbf{standalone} ES, PSO, and SA for the Ackley function}
  \label{fig:explore}
\end{figure}

To demonstrate PESA exploration capabilities, Figure \ref{fig:explore} plots the Ackley fitness mean and $1-\sigma$ standard deviation for PESA against standalone ES, PSO, and SA. The statistics are calculated based on ``all'' individuals within each generation (i.e., low and high quality solutions). Obviously, PESA features a much bigger error bar than the standalone algorithms, which reflects sample diversity within each PESA generation. The diversity of samples helps PESA to have better exploration of the search space, and overall better performance. On the other hand, the very small error bar of ES in Figure \ref{fig:explore} implies more exploitative behavior than PSO, which fairly explores the space.

\begin{figure}[h] 
 \centering
  \includegraphics[width=0.55\textwidth]{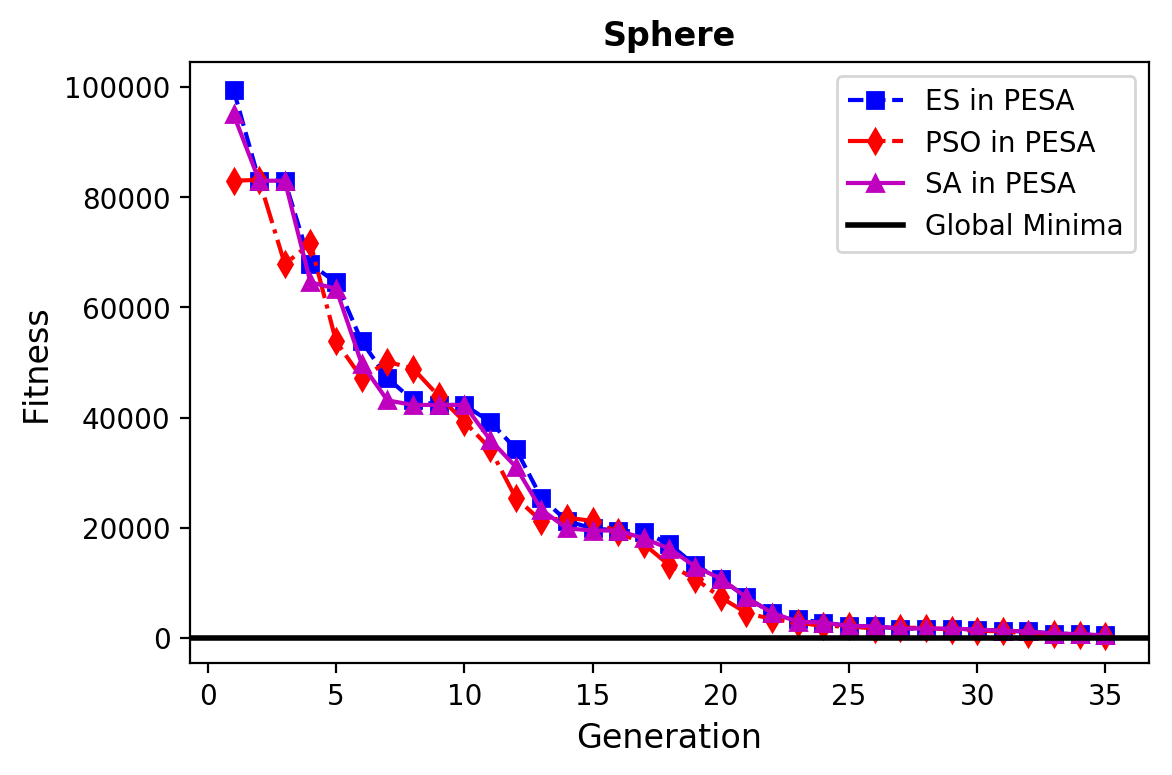}
  \caption{Fitness convergence of ES, PSO, and SA \textit{\textbf{within PESA}} for the Sphere function}
  \label{fig:prog}
\end{figure}

To explain how PESA improved the performance of the three individual algorithms, Figure \ref{fig:prog} shows how ES, PSO, and SA perform \textbf{during PESA search} (i.e. NOT as standalone algorithms). First, we notice that PSO is leading PESA search early on as can be seen from the lower PSO fitness. Afterward, experience replay takes over by first guiding SA and preventing it from divergence as seen in Figure \ref{fig:bench}, and second by allowing ES and SA to lead PESA search as can be observed in some generations between 5-10. Additionally, thanks to the backdoor greedy replay, which allows SA to stay close to ES and PSO over the whole search period, investigating relevant search regions. At the end of the evolution, the three algorithms converge to each other due to the experience share of high quality solutions across the three algorithms. Unlike Figure \ref{fig:explore}, which shows PESA exploration ability, Figure \ref{fig:prog} shows how PESA uses SA to continuously exploit toward best solutions over the whole evolution period.

%Unfortunately, the previous advantages we observed are not free, but come with additional computational costs (which should be expected), consumed mainly by 

Comparison of computational time between the four algorithms for three selected functions is given in Table \ref{tab:time}. The memory management phase in the forms of sorting the memory, updating and re-sampling from the memory, calculating the priorities, annealing $\alpha$, and cleaning the memory from duplicates result in a longer run time for PESA compared to the standalone algorithms as expected. It is important mentioning that the numbers in Table \ref{tab:time} involve running PESA and other algorithms in serial (single core), meaning that PSO, ES, and SA parts in PESA are not running in parallel, since we found that parallelization not necessary for these benchmarks. Therefore, from Table \ref{tab:time}, out of about $\sim$12s of PESA computing time, about 65\% ($\sim$7-8s) of the time is taken by the algorithms, while the rest are taken by the memory. The user can reduce the memory effect by controlling the full capacity of the memory ($D_{max}$), as interpreting large memories later in the evolution is consuming more time. The tests below in Table \ref{tab:time} involve registering every possible unique solution without limits (to obtain best performance), which would justify the significant portion consumed by the memory.  

% Table generated by Excel2LaTeX from sheet 'comptime'
\begin{table}[htbp]
  \centering
  \footnotesize
  \caption{Computing time in seconds required by different algorithms to run 100 generations for three selected functions}
    \begin{tabular}{llll}
    \toprule
    Method & Cigar & Sphere & Ackley \\
    \midrule
    PESA (serial)  & 11.7s  & 11.0s  & 12.8s \\
    ES (serial)   & 1.2s  & 1.3s   & 1.2s \\
    PSO (serial)   & 5.6s   & 5.2s   & 6.2s \\
    SA (serial)    & 0.5s   & 0.5s   & 0.9s \\
    \bottomrule
    \end{tabular}%
  \label{tab:time}%
\end{table}%

\section{Conclusions}

The concepts of experience share and replay are demonstrated through the proposed PESA algorithm to improve the search performance of evolutionary/stochastic algorithms. Experience share is preformed through connecting particle swarm optimisation (PSO), evolution strategies (ES), and simulated annealing (SA) with a replay memory, storing all their observed solutions. Experience replay is conducted by re-sampling with priority coefficient from the memory to guide the learning of all algorithms. In addition, greedy replay is used in backdoor form with SA to improve PESA exploitation behavior. The validation against 12 high-dimensional continuous benchmark functions shows superior performance by PESA against standalone ES, PSO, and SA, under similar initial starting points, hyperparameters, and number of generations. PESA shows much better exploration behaviour, faster convergence, and ability to find the global optima compared to its standalone counterparts. Given the promising performance, the authors are now focusing on fully-parallelizing PESA such that ES, PSO, and SA can evaluate each generation in shorter time. This is especially important when the fitness evaluation is complex (e.g., requires computer simulation). Additionally, PESA will go through additional benchmarking against other hybrid evolutionary methods in the literature, e.g. ES/SA or RL/PSO. Lastly, combinatorial PESA version will be developed and benchmarked in solving engineering combinatorial problems with heavy constraints.        

\section*{Acknowledgment}
This work is sponsored by Exelon Corporation, a nuclear electric power generation company, under the award (40008739)

\section*{Data Availability}
The PESA GitHub repository will be released to the public once the peer-review process is done, which will include the source implementation and wide range of unit tests from benchmarks to engineering applications.  

%\section*{References}
\bibliographystyle{elsarticle-num}
%\bibliographystyle{apa}
%\biboptions{authoryear}
{
\footnotesize \bibliography{references}}.

\begin{thebibliography}{10}
\expandafter\ifx\csname url\endcsname\relax
  \def\url#1{\texttt{#1}}\fi
\expandafter\ifx\csname urlprefix\endcsname\relax\def\urlprefix{URL }\fi
\expandafter\ifx\csname href\endcsname\relax
  \def\href#1#2{#2} \def\path#1{#1}\fi

\bibitem{abraham2006swarm}
A.~Abraham, H.~Guo, H.~Liu, Swarm intelligence: foundations, perspectives and
  applications, in: Swarm intelligent systems, Springer, 2006, pp. 3--25.

\bibitem{sanchez2012industrial}
E.~Sanchez, G.~Squillero, A.~Tonda, Industrial applications of evolutionary
  algorithms.

\bibitem{freitas2002review}
A.~A. Freitas, A review of evolutionary algorithms for e-commerce, in:
  E-Commerce and Intelligent Methods, Springer, 2002, pp. 159--179.

\bibitem{grosan2007hybrid}
C.~Grosan, A.~Abraham, Hybrid evolutionary algorithms: methodologies,
  architectures, and reviews, in: Hybrid evolutionary algorithms, Springer,
  2007, pp. 1--17.

\bibitem{kao2008hybrid}
Y.-T. Kao, E.~Zahara, A hybrid genetic algorithm and particle swarm
  optimization for multimodal functions, Applied soft computing 8~(2) (2008)
  849--857.

\bibitem{chen2009hybrid}
P.-H. Chen, S.~M. Shahandashti, Hybrid of genetic algorithm and simulated
  annealing for multiple project scheduling with multiple resource constraints,
  Automation in Construction 18~(4) (2009) 434--443.

\bibitem{ma2014hybrid}
P.~C. Ma, F.~Tao, Y.~L. Liu, L.~Zhang, H.~X. Lu, Z.~Ding, A hybrid particle
  swarm optimization and simulated annealing algorithm for job-shop scheduling,
  in: 2014 IEEE International Conference on Automation Science and Engineering
  (CASE), IEEE, 2014, pp. 125--130.

\bibitem{shen2008hybrid}
Q.~Shen, W.-M. Shi, W.~Kong, Hybrid particle swarm optimization and tabu search
  approach for selecting genes for tumor classification using gene expression
  data, Computational Biology and Chemistry 32~(1) (2008) 53--60.

\bibitem{jamasb2019novel}
A.~Jamasb, S.-H. Motavalli-Anbaran, K.~Ghasemi, A novel hybrid algorithm of
  particle swarm optimization and evolution strategies for geophysical
  non-linear inverse problems, Pure and Applied Geophysics 176~(4) (2019)
  1601--1613.

\bibitem{zameer2014core}
A.~Zameer, S.~M. Mirza, N.~M. Mirza, Core loading pattern optimization of a
  typical two-loop 300 mwe pwr using simulated annealing (sa), novel crossover
  genetic algorithms (ga) and hybrid ga (sa) schemes, Annals of Nuclear Energy
  65 (2014) 122--131.

\bibitem{rogers2009optimization}
T.~Rogers, J.~Ragusa, S.~Schultz, R.~S. Clair, Optimization of pwr fuel
  assembly radial enrichment and burnable poison location based on adaptive
  simulated annealing, Nuclear Engineering and Design 239~(6) (2009)
  1019--1029.

\bibitem{de2009particle}
A.~A. de~Moura~Meneses, M.~D. Machado, R.~Schirru, Particle swarm optimization
  applied to the nuclear reload problem of a pressurized water reactor,
  Progress in Nuclear Energy 51~(2) (2009) 319--326.

\bibitem{kropaczek1991core}
D.~J. Kropaczek, P.~J. Turinsky, In-core nuclear fuel management optimization
  for pressurized water reactors utilizing simulated annealing, Nuclear
  Technology 95~(1) (1991) 9--32.

\bibitem{mnih2015human}
V.~Mnih, K.~Kavukcuoglu, D.~Silver, A.~A. Rusu, J.~Veness, M.~G. Bellemare,
  A.~Graves, M.~Riedmiller, A.~K. Fidjeland, G.~Ostrovski, et~al., Human-level
  control through deep reinforcement learning, nature 518~(7540) (2015)
  529--533.

\bibitem{schaul2015prioritized}
T.~Schaul, J.~Quan, I.~Antonoglou, D.~Silver, Prioritized experience replay,
  arXiv preprint arXiv:1511.05952.

\bibitem{kennedy1995particle}
J.~Kennedy, R.~Eberhart, Particle swarm optimization, in: Proceedings of
  ICNN'95-International Conference on Neural Networks, Vol.~4, IEEE, 1995, pp.
  1942--1948.

\bibitem{beyer2002evolution}
H.-G. Beyer, H.-P. Schwefel, Evolution strategies--a comprehensive
  introduction, Natural computing 1~(1) (2002) 3--52.

\bibitem{kirkpatrick1983optimization}
S.~Kirkpatrick, C.~D. Gelatt, M.~P. Vecchi, Optimization by simulated
  annealing, science 220~(4598) (1983) 671--680.

\bibitem{clerc2002particle}
M.~Clerc, J.~Kennedy, The particle swarm-explosion, stability, and convergence
  in a multidimensional complex space, IEEE transactions on Evolutionary
  Computation 6~(1) (2002) 58--73.

\bibitem{jamil2013literature}
M.~Jamil, X.-S. Yang, A literature survey of benchmark functions for global
  optimisation problems, International Journal of Mathematical Modelling and
  Numerical Optimisation 4~(2) (2013) 150--194.

\end{thebibliography}

\end{document}